\documentclass[10pt,twocolumn,letterpaper]{article}

\usepackage{iccv}
\usepackage{times}
\usepackage{epsfig}
\usepackage{graphicx}
\usepackage{amsmath}
\usepackage{amssymb}
\usepackage{subcaption}

\usepackage[pagebackref=true,breaklinks=true,letterpaper=true,colorlinks,bookmarks=false]{hyperref}

\iccvfinalcopy 

\def\iccvPaperID{6703} 

\ificcvfinal\fi
\begin{document}

\title{SAWNet: A Spatially Aware Deep Neural Network for 3D Point Cloud Processing}

\author{Chaitanya Kaul, Nick Pears, Suresh Manandhar\\
Department of Computer Science, University of York\\
Heslington, York, UK, YO10 5DD\\
{\tt\small \{ck807,nick.pears,suresh.manandhar\}@york.ac.uk}
}

\maketitle

\begin{abstract}
Deep neural networks have established themselves as the state-of-the-art methodology in almost all computer vision tasks to date. But their application to processing data lying on non-Euclidean domains is still a very active area of research. One such area is the analysis of point cloud data which poses a challenge due to its lack of order. Many recent techniques have been proposed, spearheaded by the PointNet architecture. These techniques use either global or local information from the point clouds to extract a latent representation for the points, which is then used for the task at hand (classification/segmentation). In our work, we introduce a neural network layer that combines both global and local information to produce better embeddings of these points. We enhance our architecture with residual connections, to pass information between the layers, which also makes the network easier to train. We achieve state-of-the-art results on the ModelNet40 dataset with our architecture, and our results are also highly competitive with the state-of-the-art on the ShapeNet part segmentation dataset and the indoor scene segmentation dataset. We plan to open source our pre-trained models on github to encourage the research community to test our networks on their data, or simply use them for benchmarking purposes.
\end{abstract}

\section{Introduction}

Deep Learning on point clouds has progressed at a fast rate since the introduction of the PointNet \cite{pointnet} architecture. The era of Deep Learning has seen deep neural networks applied to almost every computer vision task with researchers finding innovative ideas to create novel techniques, and building on existing ones. One such field is where deep neural networks process an orderless arrangement of data points to produce semantically meaningful results. 
This sort of problem is a subset of an emerging area of deep learning, known as geometric deep learning. Such an orderless arrangement of data points is obtained from sensors which sample points of a 3D surface, resulting in a 3D point cloud. These point clouds are the closest representation to raw sensor data that is available and processing them directly in this form has greatly interested researchers, giving the created algorithms a more "end-to-end" feel, which is one of the great successes of deep neural networks. \\ 

Due to the lack of any specific arrangement available with points lying in a D dimensional space, the network needs to be invariant to their reorderings. The first papers to point this out as a potential research problem were PointNet \cite{pointnet} and Deep Sets \cite{deepsets}. They use permutation invariant and permutation equivariant functions respectively to process the points to map the data to a symmetric function, which results in the desired representation of the data. Further research \cite{pointnet2} has taken the idea of PointNet forward and applied it to various domains. Locality information has also been added to the architecture \cite{dgcnn}, where networks look at a local region in space rather than the entire point cloud in terms of its absolute coordinates. \\

We address the notion of looking both globally, as well as locally at the input points to extract meaningful features from them. We learn meaningful global point cloud embeddings using the shared Multi Layer Perceptron (MLP) to obtain a symmetric function over the entire point cloud, making it invariant to any permutations in the input. We combine this representation with the dynamic locality information from Dynamic Graph CNNs (DGCNNs) \cite{dgcnn} to create a layerwise representation for the input points that contains both global, as well as local information. We use residual identity mappings \cite{Resnet} to transfer information from earlier embeddings to later ones in the network, and to also transfer information within a layer (See \ref{sawlayer}). These connections also lead to better gradient backpropagation through the network and lead to stable learning. \\

Our contributions are as follows,
\begin{enumerate}
  \item We combine global information from one point set embedding with local information embedding for the same point set to create a hybrid layer that is "spatially aware" of its surroundings, along with having global context. 
  \item We learn residual point set embeddings, rather than the conventional (sequential) ones, which results in better accuracy over various point cloud processing tasks.
  \item We experiment with different point cloud embedding methods and provide our analysis of them. 
  \item We provide an evaluation of our network to empirically show how we arrived at this architecture.
  \item We provide results for our network on benchmark point cloud classification and segmentation datasets.
\end{enumerate}

The rest of the paper is organised as follows. We summarise the relevant literature in section \ref{relatedw}. Section \ref{method} describes our architecture. The evaluations and experiments are detailed in section \ref{eval}. We finally conclude with Section \ref{conc}. 

\section{Related Work}
\label{relatedw}

\begin{figure*}[htb]
\begin{center}
\includegraphics[width=0.9\linewidth]{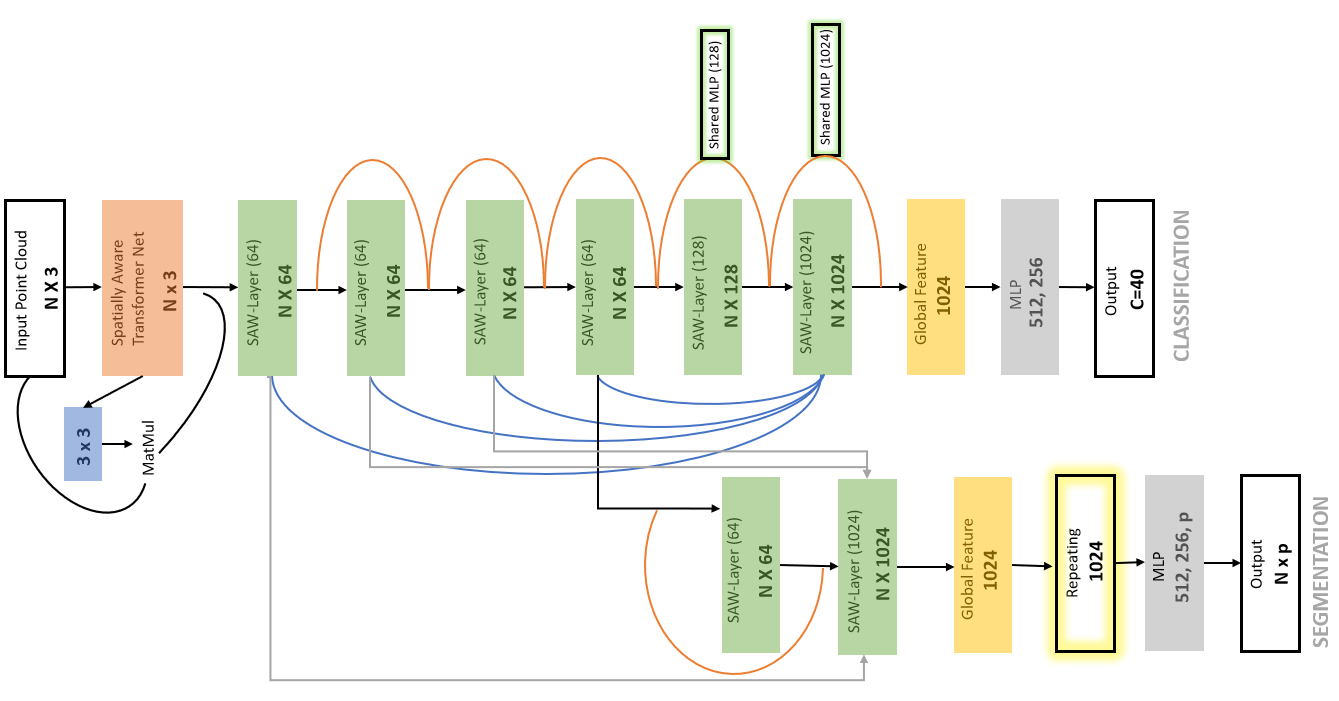}
\end{center}
\caption{Architecture for the classification and segmentation tasks. Our networks contain a spatially-aware input transformer that uses SAW-layers to regress a 3x3 transformation matrix to align the point clouds. Residual connections are used throughout to learn the point embeddings. A final global information aggregation is calculated via max pooling. The feature is then fed into a 3 layer MLP for classification. For segmentation, a $N \times p$ matrix is predicted, which is a point-wise prediction. The flow of information is from left-to-right. The orange arrows denote residual connections. The blue and grey arrows denote concatenation. Black arrows denote sequential information flow. The shared MLPs are used with the residual connections to match the embedding dimensions for adding the identity mapping.}
\label{final_arch}
\end{figure*}

\textbf{Manual Feature Extraction for 3D Shape Analysis:} Most preliminary work in 3D shape analysis approached the problem by first extracting hand-crafted features from 3D objects, and then selecting a machine learning algorithm. Usually, some local 3D shape descriptor such as 3D SIFT \cite{3dsift}, or spin image \cite{spinimage}, was used to extract features, and then they were classified using techniques such as Support Vector Machines (SVMs). Curvature-based methods \cite{curvature} have also proven successful for the task of 3D shape analysis where principal curvatures are extracted from the shape. These values can be used to calculate shape based metrics such as shape index and curvedness which encodes local shape information, giving local geometric properties of the shape. \\

\textbf{Convolutional Neural Networks:} 
Alexnet \cite{alexnet} and VGGNet \cite{vggnet} started the deep learning revolution with their state-of-the-art results on the Imagenet dataset. ResNets \cite{Resnet} hypothesised that it is easier for a network to learn a residual mapping rather than a sequential mapping of convolutions from the input to the output. ResNets are easier to train due to efficient backpropagation of gradients and they also facilitate building deeper neural nets whose accuracy doesn't degrade with increase in the number of layers. Dropout \cite{dropout} was introduced as a technique to deal with overfitting where a number of neurons, based on the dropout rate, were dropped during the computation of the forward and backward pass of the gradient. This forced all the parameters to learn useful information uniformally. Batch Normalization \cite{bn} was introduced to address the vanishing gradient problem in deeper neural network. It helps in learning stable neural networks by scaling the activation function. \\

\textbf{Deep Learning on 3D Point Clouds:} For 3D points, there is no grid-like structure to apply the convolution operation on. So several earlier techniques took the 3D points and projected them on a 2D grid like structure, following which, 2D CNNs were used on the projections. Networks such as MVCNN \cite{mvcnn} sample 80 views from the point clouds and project them to 2D, which are then processed by a CNN. This is followed by a view-pooling operation, the output of which was fed into another CNN which is used for classification. Another grid-like representation is the voxel based one. VoxNet \cite{voxnet} sample 3D points into a $32 \times 32 \times 32$ 3D occupancy grid which is processed by a 3D CNN. The networks that require projections have some notable drawbacks, the most important of which is the computational expense. Storage and processing of multiple views or voxel based data is computationally expensive and projection of the data on to a view point or a small 3D occupancy grid leads to the loss of important information. \\

The first works in processing the point clouds directly were based on the observation that the network layers need to learn an order-invariant representation of the input points. Two approaches were introduced in this regard, namely permutation invariance and permutation equivariance. The permutation invariance approach was introduced in the PointNet \cite{pointnet} paper, where the authors hypothesised learning a symmetric function to account for all $M$! permutations of the input points of a point cloud. The major drawback of this approach was that the network only looked at the global point coordinates so this representation was not as robust. Locality information was introduced by the authors in PointNet++ \cite{pointnet2} where they used farthest point sampling to uniformly sample points to build a local region, and grouped them based on a radius-based ball query to define a local region. The PointNet architecture was the applied to this local space. These networks have found applications in exploring spatial context for scene segmentation \cite{spatialcontext}, 3D object detection from RGB-D data \cite{frustrum}, and even normal and curvature estimation \cite{pcpnet} from noisy point clouds. Networks such as SO-Net \cite{sonet} perform hierarchical feature extraction using Self-Organising Maps (SOMs). The networks discussed so far capture local geometric information using local points. Dynamic Graph Convolutional Neural Networks (DGCNNs) \cite{dgcnn} compute edge features based on these points to learn a relationship between the points and its features. These networks recompute the graph based on the nearest neighbours at every layer, and hence they are termed as dynamic. Networks such as SpiderCNN \cite{spidercnn} (SpiderConv) and PointCNN \cite{pointcnn} (X-Conv) define convolution operations on point clouds based on local geometric information, rather than using a symmetric function to process them. 
\\

\textbf{Connection to graph convolutions:} Another set of closely related CNN architectures assume point clouds to be graphs, and define convolutions on these graphs by converting the points to a different domain. Keeping with the theme of global and local point processing, graph convolutions, generally, first transform the points using a fourier transform, and then learn the parameters of the convolution filters in the transformed space. This operation can be done patch wise \cite{firstgraphpaper}, or in a global sense \cite{kipf}. The transformation bases are estimated in these networks via an eigen decomposition of the graph laplacian matrix, which is a computationally expensive process. Further, computing convolution filters in the transformed space also has added computational complexity, though techniques such as \cite{fastergcnn} localise convolution filters faster. \\ 

We extend the ideas of PointNet and DGCNNs to learn both local and global information for points in every layer. This makes our approach robust to problems such as occlusions, while not losing the `global awareness' of the point cloud's space in terms of its absolute coordinates.

\section{Methodology}
\label{method}

In this section, we explain the network architecture for Spatially AWare Network.

\subsection{Architecture Overview}

The architecture takes a $N \times 3$ point cloud as input and outputs a class label or a per point semantic segmentation label for it. The input is fed into a transformer net that uses SAW-Layers to regress a transformation matrix for point cloud alignment. The aligned points are then fed into a series of SAW-Layers to create a permutation invariant embedding of the points. We use residual connections to transfer information between layers. This allows us to train deeper and more stable network architectures. For layers where the embedding size increases, the residual mappings account for that with a shared MLP before the addition of the previous embedding with the next. We concatenate the output of every previous SAW-Layer before passing it into the final 1024-D SAW-Layer to get a better global contextual symmetry aggregation. The output of the final 1024-D embedding is maxpooled over, to get a single 1024 dimensional vector, which contains the global contextual information for the point clouds. This information is used to classify or segment the points using a MLP. For the segmentation tasks, the output of the MLP is reshaped to get a per point semantic segmentation.

\subsection{Spatially Aware Transformer Net}

The spatially aware transformer network contains 3 SAW-Layers which embed the point cloud into a 64, 128 and a 1024 dimensional space respectively. The output embedding from the final layer is aggregated into a global context (pooled) feature vector that is fed into a multi-layer perceptron with two hidden layers of size 512 and 256. The output is a final dense layer of size 9 that is reshaped into a $3 \times 3$ transformation matrix, the elements of which are the learnt affine transformation values for the point cloud. A simple matrix multiplication of this matrix with the input point cloud, aligns the points in 3D space. The architecture can be seen as a mini version of the SAW-Net architecture.

\subsection{Spatially AWare-Layers}
\label{sawlayer}

\begin{figure}[t]
\begin{center}
\includegraphics[width=0.7\linewidth, scale=1.0]{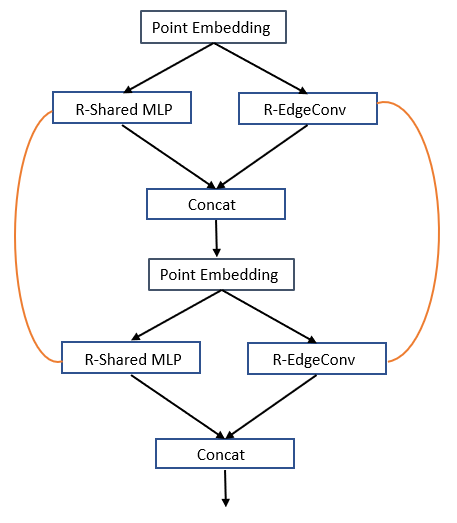}
\end{center}
\caption{The figure shows how information propagates in our network between two SAW-Layers. There are residual connections for learning the individual local and global embeddings within the layer, as well as information propagation to the next layer via a skip connection. This forces the local layer to keep focusing on the local properties and the global layer to keep focusing on the global one.}
\label{singlelayer}
\end{figure}

The architecture of a single SAW-Layer can be seen in figure \ref{singlelayer}. Consider the input to the layer to be the point embedding from an intermediate layer. This input is then examined globally by the R-shared MLP layer, and also by computing a dynamic graph, based on 20 nearest neighbors, and then learning the dependency of the neighbours on the point, using the R-edgeconv layer. The output from the R-edgeconv layer is then maxpooled across the neighbors, concatenated with the output from the R-shared MLP, and fed into the next layer. Each of the R-shared MLP and R-edgeconv layers individually contain two shared MLPs within them. The first R-shared MLP computes an embedding and has a batch normalization and ReLU computation as a part of its output estimation. The second shared MLP within it has a batch normalization operation and then adds the input to the output, before passing it through an activation function. The R-edgeconv has a similar residual mapping within it. The two layers also individually transfer information to the next embedding layer using residual connections. \\

Consider a set of points $X = x_1, ... x_n$, where $x_n \in R^D$. So each point is a D dimensional embedding of some point cloud. These points are processed by the SAW-Layer in the following way. The input is passed in parallel into the R-shared MLP and the R-edgeconv layers. The R-shared MLP first applies a transformation on the points, $f(t) = f(x_1, ... x_n) = \{h_1(x_1),...,h_1(x_n)\}, \ h: R^D \to R^M $ where M is the length of the embedding. $h_1$ here is the shared weighted MLP. The output of the first shared MLP layer is $S_1 = \sigma(B_1(f(t)))$. $\sigma$ is the activation function ReLU and $B_i$ is the batch normalization for the $i^{th}$ instance. This output is then fed into another shared MLP - BN combination. The output can be represented as, $S_2 = B_2(h_2(S_1))$. An identity mapping is then added to this output, $G = S_2 + X$ giving us the output of the R-shared MLP, G, which is the global feature computation. Similarly, if $e$ denotes the edgeconv operation, which computes the dependence of a point on its k - nearest neighbours, and $f(k)$ denotes the input in terms of its k nearest neighbours, called the edge feature, then one pass of the input $X$ through edgeconv-BN-ReLU would give $E_1 = \sigma(B_r(e_1(f(k))))$. Applying another edgeconv-BN would give $E_2 = B_s(e_2(E_1))$. This value is then maxpooled over its nearest neighbours, following which an identity mapping is added to the output $E_2$. This can be shown as $L = E_2 + X$, where $L$ is the computed local feature. These values are the same in dimension, and are concatenated together, to form the output of the SAW-Layer, $SAW-Layer = [G:L]$, where $[\ :\ ]$ denotes concatenation across the $n^{th}$ dimension of a tensor of size $T^{u \times v \times w \times n}$. This point embedding is fed into the subsequent SAW-Layers.

\section{Evaluation}
\label{eval}

We test the performance of our network on point cloud classification and point cloud segmentation tasks.

\subsection{Classification}

We use the ModelNet40 dataset \cite{shapenet} which contains 12,311 CAD models for 40 objects in 3D. The models are man made for the 40 object categories and the dataset is divided into a training and test split of 9,843 and 2,468 respectively. We use the same data split for our experiments and report the results on the test set. \\

To demonstrate that our method improves performance over PointNet and DGCNN, we use the same pre-processing as in their experiments. Hence, for each of the 3D models, we sample 1024 points uniformly from the mesh faces and normalise these points to a unit sphere. The dataset contains points as well as surface normals. We only consider the point cloud coordinates and discard the remaining information for our experiments. During training, we use data augmentation to add variations in our dataset. Random rotations and scaling is added along with per point jitter to perturb the location of the points. This is the same strategy as was used for data augmentation in PointNet++ training. \\

\subsubsection{Training}
\label{train}

We use Tensorflow \cite{tensorflow} for all our implementations for this paper, unless explicitly stated otherwise. For the classification task, we use the same training setting as in \cite{pointnet}. Our classification network was trained using a batch size of 8, using one Nvidia GTX 1080 Ti and categorical cross entropy loss. We use Adam \cite{adam} with a learning rate of 0.001, which is halved after every 20 epochs. The decay rate for batch normalization is 0.7. We train our model for 250 epochs.  

\subsubsection{Results}

\begin{table}
\begin{center}
\begin{tabular}{p{3.8cm} p{1.5cm} p{1.5cm}}
\hline
Method & Class Accuracy (\%) & Instance Accuracy (\%)\\
\hline\hline
3D ShapeNets \cite{shapenet} & 77.3 & 84.7\\
VoxNet \cite{voxnet} & 83.0 & 85.9\\
Subvolume \cite{subvolume} & 86.0 & 89.2\\
ECC \cite{ECC} & 83.2 & 87.4\\
PointNet \cite{pointnet} & 86.0 & 89.2\\
PointNet++ \cite{pointnet2} & - & 90.7\\
KD-Net (Depth 10) \cite{kdnet} & 86.3 & 90.6\\
KD-Net (Depth 15) \cite{kdnet} & 88.5 & \textbf{91.8}\\
DGCNN \cite{dgcnn} & 89.2 (90.2)* & 91.6 (92.2)*\\
SO-Net \cite{sonet} ($2048 \times 3$)& 87.3 & 90.9\\
SpiderCNN \cite{spidercnn} ($1024 \times 3$)& - & 90.5\\
\hline
Ours & \textbf{90.0} & \textbf{91.8}\\
\hline
\end{tabular}
\end{center}
\caption{Classification results on the ModelNet40 dataset. The asterisk (*) denotes the results presented in the paper. Quoted result is based on our re-runs of the author's github repository.}
\label{ModelNet40}
\end{table}

Our results are summarised in Table \ref{ModelNet40}. The table shows the results for all architectures with their classification results using the (x,y,z) coordinates. All networks shown in the table are trained on 1024 points except SO-Net, which takes twice the number of points as its input. Our network performs considerably well with respect to the current state of the art architectures. It has the best average accuracy, and the joint-best average class accuracy among the networks.

\subsection{Further Experiments on ModelNet40}

We used the ModelNet40 dataset for a series of tests to finalise our network architecture. 

\subsubsection{Experiments with PointNet}

\begin{table}
\begin{center}
\begin{tabular}{p{3.8cm} p{3cm}}
\hline
Method & Accuracy (\%) \\
\hline\hline
PointNet & 88.84\\
Grouped Embeddings & 89.3\\
Depthwise Embeddings & 89.4\\
Residual Embeddings & 90.0\\
\hline
\end{tabular}
\end{center}
\caption{Experimentation with learning different embeddings for the PointNet architecture. Results are based on our re-implementation of the architecture with the Keras API on the ModelNet40 dataset. The reported value is the instance accuracy.}
\end{table}

We re-implemented the PointNet architecture with the aim of testing the effectiveness of the shared MLP against other methods for embedding learning. All networks for these experiments were implemented in Keras \cite{keras}. \\

For the first experiment, we replaced the shared MLP layer with a grouped \cite{grouped} MLP layer, where we split the MLP into different groups within the same layer, which forces each group to learn to give outputs independent of the other groups. This can be seen as learning multiple shared MLPs within a single layer. Weights are shared within the groups. In the next experiment, the shared MLP layer was replaced by a depthwise \cite{depthwise} shared MLP layer. Here, we learn individual functions for each point in the point cloud, and then use a 
weight-shared MLP to learn a symmetric function over those functions. Our final experiment consisted of using a residual mapping with the conventional shared MLP. This method gave the best performance increase with PointNet and so we embedded this methodology into our final architecture. It is useful to note that the grouped MLP and the depthwise shared MLP did offer performance increases over the current shared MLP layer as well, but this increase in accuracy came with its own addition to computational complexity. The residual mapping is chosen mainly due to its greater performance increase, as well as its ease of integration into existing architectures.

\subsubsection{Ablation Tests}

\begin{figure}
\centering
\begin{subfigure}[b]{0.45\textwidth}
  \centering
  \includegraphics[width=1.0\linewidth]{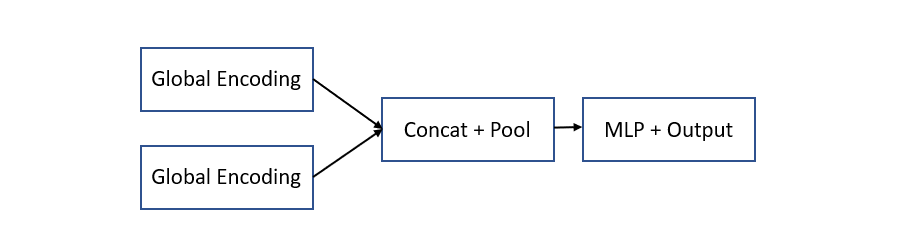}
  \caption{}
  \label{ablation1}
\end{subfigure}
\\
\begin{subfigure}[b]{0.45\textwidth}
  \centering
  \includegraphics[width=1.0\linewidth]{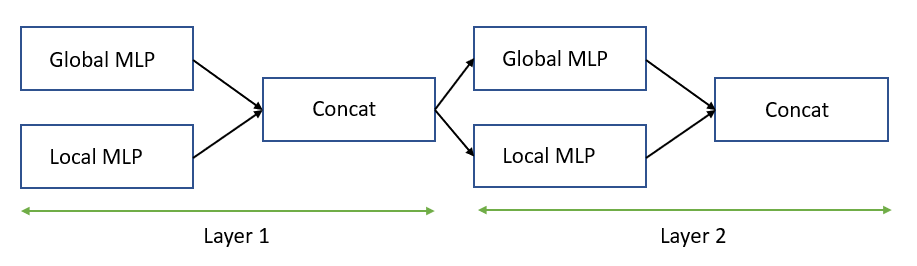}
  \caption{}
  \label{ablation2}
\end{subfigure}
\\
\begin{subfigure}[b]{0.45\textwidth}
  \centering
  \includegraphics[width=1.0\linewidth]{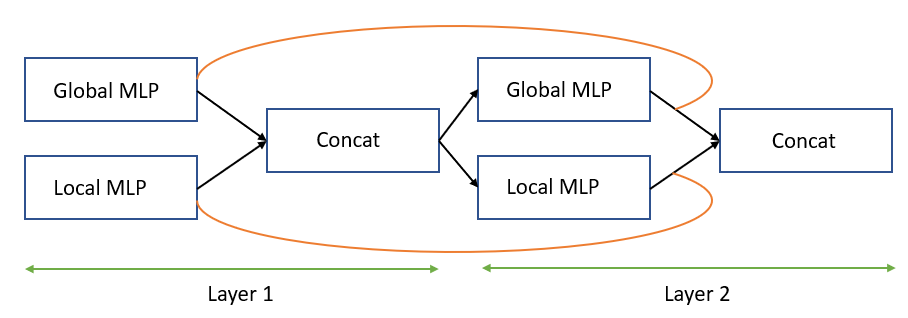}
  \caption{}
  \label{ablation3}
\end{subfigure}
\caption{Different ablation setups experimented with.}
\label{ablation}
\end{figure}

\begin{table}
\begin{center}
\begin{tabular}{l c}
\hline
Method & Accuracy (\%) \\
\hline\hline
Combine at end & 88.9\\
Combine per layer & 89.6\\
Residual + Combine per layer & 90.0\\
Residual + Combine per layer + 1 FC & 63.6\\
\hline
\end{tabular}
\end{center}
\caption{Ablation tests on the architecture. Results are the values for the class accuracy on the ModelNet40 dataset.}
\end{table}

Certain combinations of local and global geometric properties work better than others in terms of performance. We combined the global and local vectors at different stages in the network to test our hypotheses. Our first experiment comprised of running two parallel blocks of embeddings, each looking at global (PointNet), and local (DGCNN) geometric properties respectively. Global feature aggregation over the outputs of these embeddings was concatenated to give a vector that comprised of information from both networks. This was then fed into a 3 layer MLP for classification. We also computed, and then combined, the global and local feature vectors per layer, and computed further embeddings based on this combination. \\

The combination per layer results in a better accuracy than combining the features at the end. This adds spatial awareness to the network, which it further uses to compute a better embedding of the points. \\

We then added residual connections to this setting which further improved the performance of the networks by 0.4\% and we used this as our final architecture. Further experiments were conducted, such as removing the fully connected layers, but they all resulted in a drastic drop in the accuracy of the network. This leads us to believe that the fully connected layers are an important part of the architecture.  

\subsubsection{Robustness to missing points} 

\begin{figure}[t]
\begin{center}
\includegraphics[width=1.0\linewidth]{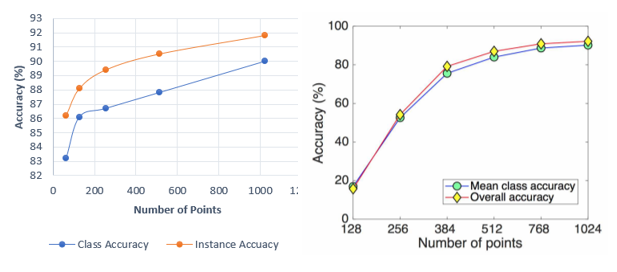}
\end{center}
   \caption{Comparing the robustness to sparse point inputs of our network with DGCNN.}
\label{Robust}
\end{figure}

To test how our network would do with sparse input points, we observed the degradation of our network's accuracy with respect to a small number of input points. We ran the network for 75 epochs with the same hyperparameters as discussed in Section \ref{train}, and recorded the best test set accuracy. Surprisingly, the accuracy of the network does not degrade to a large extent, even when the number of points input to the network drops down to 128, showing how robust our network is to random point dropout. Figure \ref{Robust} shows the comparison of our observation with DGCNN. The graph on the left is the performance of our architecture, and the graph on the right has been reproduced from the DGCNN paper \cite{dgcnn}. The performance of DGCNN reduces considerably when the number of points drop below 384, while our network still manages to maintain a 86.1\% accuracy, even with 128 points. 

\subsection{Segmentation}

\begin{table*}[htb]
\begin{center}
\begin{tabular}{p{1.5cm} | p{0.5cm} | p{0.5cm} p{0.5cm} p{0.5cm} p{0.5cm} p{0.5cm} p{0.5cm} p{0.5cm} p{0.5cm} p{0.5cm} p{0.5cm} p{0.5cm} p{0.5cm} p{0.5cm} p{0.5cm} p{0.5cm} p{0.5cm}}
\hline
Method & IoU & Aero & Bag & Cap & Car & Chair & EarP & Guit & Knif & Lamp & Lapt & MotB & Mug & Pstl & Rckt & Skte & Tbl\\
\hline\hline
PointNet \cite{pointnet} & 83.7 & \textbf{83.4} & 78.7 & 82.5 & 74.9 & 89.6 & 73.0 & \textbf{91.5} & 85.9 & 80.8 & 95.3 & 65.2 & 93.0 & 81.2 & 57.9 & 72.8 & 80.6\\
PointNet++ \cite{pointnet2} & \textbf{85.1} & 82.4 & 79.0& 87.7 & 77.3 & 90.8 & 71.8 & 91.0 & 85.9 & \textbf{83.7} & 95.3 & 71.6 & 94.1 & 81.3 & 58.7 & \textbf{76.4} & \textbf{82.6}\\
Kd-Net \cite{kdnet} & 82.3 & 80.1 & 74.6 & 74.3 & 70.3 & 88.6 & 73.5 & 90.2 & 87.2 & 81.0 & 94.9 & 57.4 & 86.7 & 78.1 & 51.8 & 69.9 & 80.3 \\
SO-Net \cite{sonet} & 84.6 & 81.9 & 83.5 & 84.8 & \textbf{78.1} & 90.8 & 72.2 & 90.1 & 83.6 & 82.3 & 95.2 & \textbf{69.3} & \textbf{94.2} & 80.0 & 51.6 & 72.1 & \textbf{82.6} \\
DGCNN \cite{dgcnn} & \textbf{85.1} & 84.2 & 83.7 & 84.4 & 77.1 & 90.9 & \textbf{78.5} & \textbf{91.5} & 87.3 & 82.9 & \textbf{96.0} & 67.8 & 93.3 & \textbf{82.6} & \textbf{59.7} & 75.5 & 82.0\\
\hline
Ours & 84.8 & 82.9 & \textbf{85.5} & \textbf{88.7} & 78.0 & \textbf{90.9} & 77.0 & 91.0 & \textbf{88.7} & 82.5 & 95.5 & 63.6 & \textbf{94.2} & 77.6 & 57.0 & 74.8 & 81.9 \\
\hline
\end{tabular}
\end{center}
\caption{Part segmentation results on the ShapeNet part dataset. The evaluation metric is mean intersection over union (mIoU).}
\label{part_segt}
\end{table*}

We now compare the performance of our network with respect to the state of the art in point cloud segmentation. For this section, the architecture is the same as the segmentation architecture shown in Figure \ref{final_arch}. We embed the transformer aligned 3D points into a higher dimensional space using SAW-Layers, and finally use a global max pooling operation to get a point cloud statistic. We use this statistic as the input into a 3 layer MLP which is then reshaped to give the point wise segmentation result. We ran this network with a batch size of 8 on 2 Nvidia GTX1080 Tis. 

\begin{figure}[htb]
\begin{center}
\includegraphics[width=0.8\linewidth]{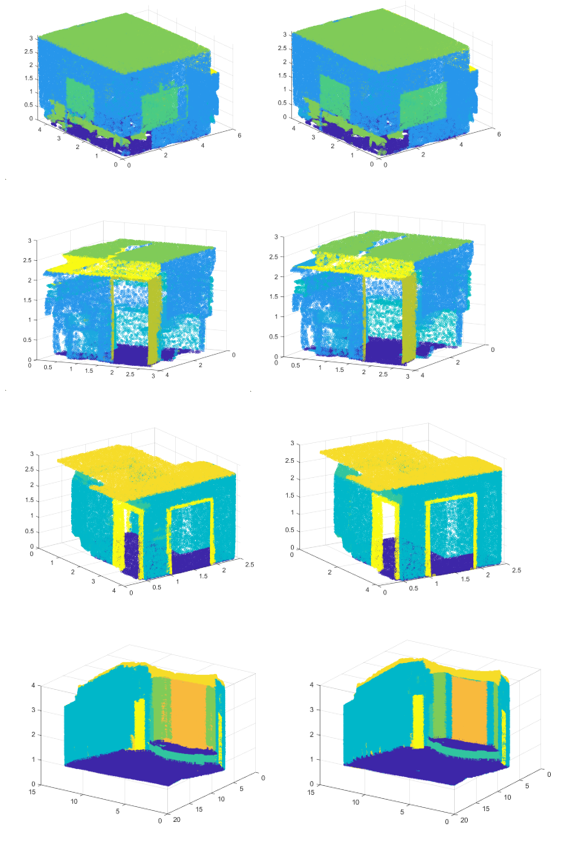}
\end{center}
\caption{Segmentation results on the Stanford semantic scene parsing dataset. The left column shows the predicted segmentation, while the right column shows the corresponding ground truths.}
\label{semanticscene}
\end{figure}

\subsubsection{3D Part Segmentation}

For this section, we use the shapenet part dataset introduced in \cite{shapenetpart}. Given a 3D point cloud, the task at hand is to segment semantic parts of the point cloud. (Eg, for an aeroplane, the wings, the body etc.). The dataset contains 16,881 3D models of 16 object categories with 50 part segmentation ground truths. Table \ref{part_segt} shows our results on the dataset. The evaluation metric for this task is the mean intersection over union (mIoU) for all the shapes in a particular category. We use the official train/val/test split for consistency with other results. \\

\begin{figure*}
\begin{center}
\includegraphics[width=0.8\linewidth]{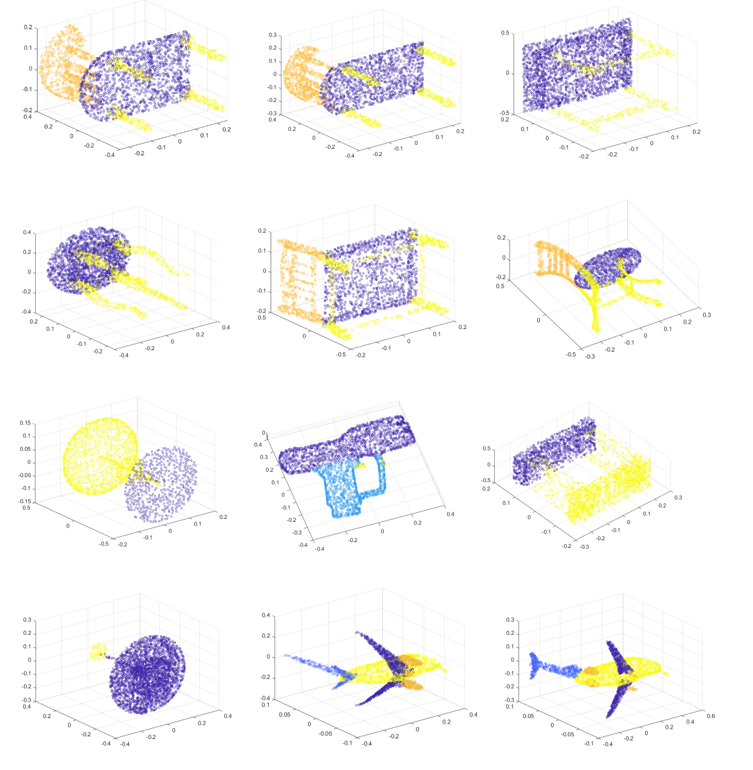}
\end{center}
\caption{Part segmentation examples from the ScanNet dataset. The samples were randomly selected for visualisation purposes.}
\label{partseg}
\end{figure*}

We get a competitive mIoU value of 84.8, which is only 0.3 less than the state of the art for this class of architectures. It is encouraging to note that even on the objects that the architecture does not get state of the art results, it mostly gets a IoU value close to the best for that object. Further, we get these results with default hyper parameters, without their explicit tuning. Some visualizations of the part segmentation output are shown in Figure \ref{partseg}. 

\subsubsection{Semantic 3D Scene Segmentation}

We also ran our network for the task of 3D semantic parsing of large scale indoor spaces using the data in \cite{sceneparsing}. We use the training/testing set up as in \cite{pointnet} with the difference that we train our model on multiple GPUs. Each input point in this dataset is a 9D vector. We use only the {X,Y,Z} coordinates to calculate the R-edgeconv embedding, and all the 9 dimensions are fed into the R-shared MLP for the input. The rest of the embeddings are calculated similar to the architecture in Figure \ref{final_arch}. All the 4096 points are used to train the model on 2 Nvidia GTX 1080 Tis with a batch size of 4 per GPU. We show quantitative results on this dataset in Figure \ref{semanticscene}.

\section{Conclusion}
\label{conc}

We proposed a deep neural network architecture that considers local and global geometry of points per network layer in order to process point clouds more efficiently. Our results show state-of-the-art performance on the ModelNet40 classification dataset, and also highly competitive performance on two point cloud segmentation datasets. Through a series of experiments to evaluate our network, we showed that, in this setting, residual learning helps to learn efficient point cloud embeddings. Our architecture is a lot more robust to random point dropout when compared to DGCNN \cite{dgcnn}. Therefore, we hypothesise that such a network may perform well in applications where occlusions are commonplace and can confound many non-robust approaches. Therefore, as future work, we aim to test the SAWNet's effectiveness on 3D face recognition tasks.

{\small
\bibliographystyle{ieee}
\bibliography{egbib}
}
\nocite{*}
\end{document}